\documentclass[final]{cvpr}

\usepackage{times}
\usepackage{epsfig}
\usepackage{graphicx}
\usepackage{amsmath}
\usepackage{amssymb}

\usepackage{booktabs}
\usepackage{multirow}
\usepackage{fixmath}
\usepackage{subcaption}
\usepackage{float}
\usepackage{xspace}

\newcommand\blfootnote[1]{%
  \begingroup
  \renewcommand\thefootnote{}\footnote{#1}%
  \addtocounter{footnote}{-1}%
  \endgroup
}

\usepackage[pagebackref=true,breaklinks=true,colorlinks,bookmarks=false]{hyperref}

\def\cvprPaperID{3422} 
\def\confYear{CVPR 2021}

\begin{document}

\title{EnD: Entangling and Disentangling deep representations for bias correction}

\author{
Enzo Tartaglione \\
University of Turin, \\
Computer Science Dept. \\
{\tt\small enzo.tartaglione@unito.it}
\and
Carlo Alberto Barbano\\
University of Turin, \\
Computer Science Dept. \\
{\tt\small carlo.barbano@unito.it}
\and
Marco Grangetto \\
University of Turin, \\
Computer Science Dept. \\
{\tt\small marco.grangetto@unito.it}
}

\maketitle

\newcommand{\absname}{EnD\xspace}

\newcommand{\country}{Italy\xspace}
\newcommand{\cdss}{Citt\'a della Salute e della Scienza\xspace}
\newcommand{\slg}{San Luigi Gonzaga\xspace}
\newcommand{\corda}{CORDA\xspace}
\newcommand{\cordacdss}{CORDA-CDSS\xspace}
\newcommand{\cordaslg}{CORDA-SLG\xspace}

\newcommand{\STAB}[1]{\begin{tabular}{@{}c@{}}#1\end{tabular}}

\begin{abstract}
Artificial neural networks perform state-of-the-art in an ever-growing number of tasks, and nowadays they are used to solve an incredibly large variety of tasks. There are problems, like the presence of biases in the training data, which question the generalization capability of these models. 
In this work we propose EnD, a regularization strategy whose aim is to prevent deep models from learning unwanted biases. In particular, we insert an ``information bottleneck'' at a certain point of the deep neural network, where we disentangle the information about the bias, still letting the useful information for the training task forward-propagating in the rest of the model. One big advantage of EnD is that we do not require additional training complexity (like decoders or extra layers in the model), since it is a regularizer directly applied on the trained model. Our experiments show that EnD effectively improves the generalization on unbiased test sets, and it can be effectively applied on real-case scenarios, like removing hidden biases in the COVID-19 detection from radiographic images.
\end{abstract}

\section{Introduction}
\label{sec:introduction}
\blfootnote{\noindent This work has been accepted as a conference paper for the 2021 Conference on Computer Vision and Pattern Recognition (CVPR 2021).}
In the last two decades artificial neural network models (ANNs) received huge interest from the research community. Nowadays, complex and even ill-posed problems can be tackled provided that one can train a deep enough ANN model with a large enough dataset. Furthermore, they aim to become a powerful tool helping us take a variety of decisions: for example, AI is currently used for scouting and hiring people~\cite{laumer2015impact}. These ANNs are trained to process a desired output from some inputs. We have no clear idea how the information is effectively processed inside. 
Recently, AI trustworthiness has been recognized as major prerequisite for people and societies to use and accept such systems~\cite{hleg2019ethics, zhang2019artificial}. In April 2019, the High-Level Expert Group on AI of the European Commission defined the three main aspects of trustworthy AI \cite{hleg2019ethics}: it should be lawful, ethical and robust. Providing a warranty on this topic is currently a matter of study and discussion.\\
Focusing on the concept of robustness for AI, Attenberg~\emph{et~al.} discussed the problem of finding the so-called ``unknown unknowns''~\cite{attenberg2015beat} in data. These unknown unknowns relate to the case when the deep model elaborates information in an unintended way, but shows high confidence on its predictions. Such behavior affected many recent works proposing AI-based solutions on the COVID detection from radiographic images. Unfortunately, the available datasets at the beginning of the pandemic were heavily biased. This often resulted in models predicting COVID diagnosis with a high confidence, thanks to the presence of unwanted biases, for example by detecting the presence of catheters or medical devices for positive patients, their age (at the beginning of the pandemic, most ill patients were elderly people), or even by recognizing the origin of the data itself (when negative cases were augmented borrowing samples from other datasets)~\cite{apostolopoulos2020covid, sethy2020detection, tartaglione2020unveiling}.\\ 
In this work we propose a regularization strategy which \underline{En}tangles the deep features extracted by patterns belonging to the same target class and \underline{D}isentangles the biased features: we name it EnD, and with it we wish to put an end 
to the bias propagation in any deep model. We assume we know data might have some bias (like in the case of COVID, the origin of data) but we ignore what it translates into (we do not have a prior knowledge on whether the bias is the presence of some color, a specific feature in the image or anything else). EnD regularizes the output of some layer $\Gamma$ within the deep model in order to create an ``information bottleneck'' where the regularizer:
\begin{itemize}
    \item entangles the feature vectors extracted from data belonging to the same target class;
    \item disentangles the features extracted from data having the same ``bias label''.
\end{itemize}
Since the deep model is trained minimizing both the loss and EnD, all the biased features are discouraged to be extracted in favor of the unbiased ones. Compared to other de-biasing techniques, we have no training overhead: we do not train extra models to perform gradient inversion on the biased information or involve the use of GaNs, or even de-bias the input data. EnD works directly on the target model, and is minimized via standard back-propagation.\\
In general, directly tackling the problem of mutual information's minimization is hard, given both its non-differentiability and the computational complexity involved. Nonetheless, previous works have already shown that adding further constraints to the learning problem could be effective~\cite{9343062} as, typically, the trained ANN models are over-sized and allows a large number of solutions to the same learning task~\cite{tartaglione2019take}. Our experiments show that EnD effectively favors the choice of unbiased features over the biased ones at training time, yielding competitive generalization capabilities compared to models trained with other un-biasing techniques.\\
The rest of the work is structured as follows. In Sec.~\ref{sec:related} we review some works close to our problem. Then, in Sec.~\ref{sec:method} we introduce EnD in detail providing intuitions on its effect. Sec.~\ref{sec:experiments} shows some empirical results and finally, in Sec.~\ref{sec:conclusion}, the conclusions are drawn. 
\section{Related works}
\label{sec:related}

In this section we review state-of-the-art techniques designed to prevent models from learning biases. The techniques can be grouped into (but not limited to) three main approaches: direct data de-biasing from the source, use of GANs/ensembling towards data de-biasing and direct learning the de-biasing within the trained model.\\ 
\noindent \textbf{De-biasing from data source} It is known that datasets are typically affected by biases. In their work, Torralba~and~Efros~\cite{torralba2011unbiased} showed how biases affect some of the most commonly used datasets, drawing considerations on the generalization performance and classification capability of the trained ANN models. Following a similar approach, Tommasi~\emph{et~al.}~\cite{tommasi2017deeper} conducted experiments reporting differences between a number of datasets and verifying how final performances vary when applying different de-biasing strategies in order to balance data. Working at the dataset level is in general a critical aspect, and greatly helps in understanding the data and its structure~\cite{Cubuk_2019_CVPR}. 
The concept of removing bias by using data borrowed by different sources has been explored in a practical and empirical context by Gupta~\emph{et~al.}~\cite{gupta2018robot}. In particular, they have designed a de-biasing strategy to minimize the effects of imperfect execution and calibration errors by reducing the effect of unbalanced data, showing improvements in the generalization of the final model.

\noindent \textbf{Adversarial and ensembling approaches.} Having an explicit formulation for the bias contribution in the loss term is typically hard. One possible approach is to use additional models to learn the biases in data and use them to condition the primary model so that it avoids them. 
Kim~\emph{et~al.} use adversarial learning and gradient inversion to eliminate the information related to the biases in the model~\cite{Kim_2019_CVPR}. 
Another possibility is to use the gray-level co-occurence matrix to extract unbiased features and to train the model, as proposed by Wang~\emph{et~al.} with HEX~\cite{wang2018learning}. Alvi~\emph{et~al.} propose the BlindEye~\cite{alvi2018turning} technique, where they train a classifier on the extracted deep features to retrieve information from biases: then, they force the ``bias classifier'' to be no longer able to retrieve bias-related information, modifying the deep features accordingly.
Bahng~\emph{et~al.}~\cite{bahng2019rebias} develop an ensembling-based technique, called \emph{ReBias}. It consists in solving a min-max problem where the target is to promote the independence between the network prediction and all biased predictions. Identifying the ``known unknowns''~\cite{attenberg2015beat} and optimize on those using a neural networks ensemble is the approach proposed by Nam~\emph{et~al.} with their LfF~\cite{nam2020learning}. A similar approach is followed by Clark~\emph{et~al.} in their LearnedMixin~\cite{ClarkYZ19}.

\noindent \textbf{De-biasing within the deep model.} Dataset de-biasing helps in the learning process, as training is performed with no biases; however, with such an approach we typically have no direct control on the information we are removing from the dataset itself, or we are including an extremely-high computational complexity like when training GANs. A context in which, on the contrary, we can have direct access to these biases is presented by Hendricks~\emph{et~al.}~\cite{hendricks2018women}. In such a work it was possible to explicitly introduce a corrective loss term (coherent with the formulation introduced by Vinyals~\emph{et~al.}~\cite{vinyals2015show}) with the aim to help the ANN model to focus on the correct features. Similarly, Cadene~\emph{et~al.} propose RUBi~\cite{cadene2019rubi} where they use logit re-weighting to lower the bias impact in the learning process, and Sagawa~\emph{et~al.}, with Group-DRO~\cite{sagawa2019distributionally}, avoid bias overfitting by defining prior data sub-groups and controlling their generalization. 
EnD belongs to this class of approaches, since we directly regularize the trained model, with no additional parameters to be learned. In Sec.~\ref{sec:method} we are going to describe in detail the approach we take in order to EnD bias propagation in the trained model.
\section{Entangling and Disentangling deep representations}
\label{sec:method}
\begin{figure}
    \centering
    \includegraphics[width=\columnwidth]{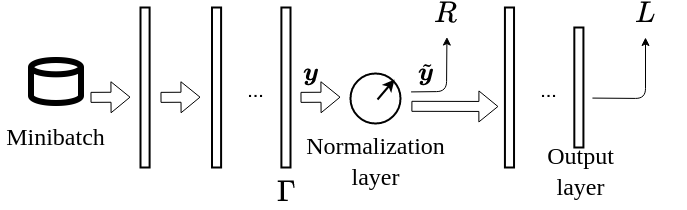}
    \caption{\textbf{Model overview.} The features for EnD are extracted at the output of $\Gamma$, after a normalization layer performing the operation as in \eqref{eq:norm}.}
    \label{fig:structure}
\end{figure}
\begin{figure}
	\centering
    \begin{minipage}{0.45\columnwidth}
        \centering
        \includegraphics[width=\textwidth]{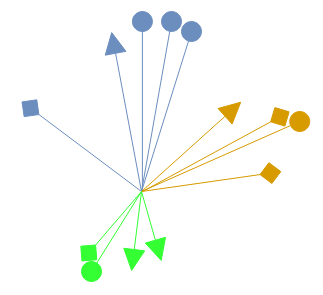}
        \caption*{(a)}
    \end{minipage}
    \begin{minipage}{0.45\columnwidth}
        \centering
        \includegraphics[width=0.85\textwidth]{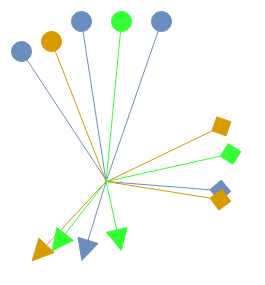} 
        \caption*{(b)}
    \end{minipage}
    \caption{\textbf{Toy example of EnD's effect.} Each arrow represents the feature vector associated with a sample. 
    Biases are represented by the three different colors (green, orange and blue) while the target class is represented by the arrows marker's symbol (triangle, square and circle). While in some un-regularized training the deep model strongly correlates with the bias (a), using EnD we aim at enforcing the choices of different features (b).}
    \label{fig:orthoeffect}
\end{figure}
In this section, after introducing the notation, we present EnD, our proposed regularization term, whose aim is to regularize the deep features in order to discourage the deep model to learn biases.

\subsection{Preliminaries}
In this section we first introduce the notation we are going to use for the rest of this work and we provide some intuitions on how EnD is going to work. Let us assume we focus our attention on some layer $\Gamma$, at the output of which we are going to apply EnD. Let $T$ be the cardinality of the target classes of the learning problem and $B$ the cardinality of the bias classes in the dataset. We say the output of $\Gamma$ is $\boldsymbol{y}~\in~\mathbb{R}^{N_{\Gamma}\times M}$, where $M$ is the batchsize and $N_{\Gamma}$ is the output size of $\Gamma$.\\
We also define:
\begin{itemize}
    \item $M^{t, b}$ as the cardinality of the samples having the same target $t$ and the same bias $b$;
    \item $M^{t, -}$ as the cardinality of the samples having the same target $t$ regardless the biases;
    \item $M^{-, b}$ as the cardinality of the samples having the same bias $b$ regardless the target class;
    \item $\boldsymbol{y}^{t,b}$ as the subset of the features $\boldsymbol{y}$ belonging to the inputs having the same target class $t$ and showing the same bias $b$;
    \item $\boldsymbol{y}^{t,-}$ as the subset of the features $\boldsymbol{y}$ belonging to the inputs having the same target class $t$ regardless the bias;
    \item $\boldsymbol{y}^{-,b}$ as the subset of the features $\boldsymbol{y}$ belonging to the inputs having the same bias $b$ regardless the target class;
    \item $\boldsymbol{y}_i$ as the $i$-th sample in the minibatch;
    \item $\mathcal{T}(\boldsymbol{y}_i)$ extracts the target class of $\boldsymbol{y}_i$;
    \item $\mathcal{B}(\boldsymbol{y}_i)$ extracts the bias class of $\boldsymbol{y}_i$.
\end{itemize}
In our work, EnD sides the loss minimization, discouraging the selection of biased deep features and encouraging the unbiased ones at training time. Hence, the overall objective function we aim to minimize is
\begin{equation}
    \label{eq:uprule}
    J = L + R,
\end{equation}
where $L$ is the loss function for the trained task and $R$ is our proposed EnD term, applied at the output of $\Gamma$. Fig.~\ref{fig:structure} provides the overall structure of the trained model.\\
Let us consider, as a toy example, some classification problem having three target classes, but as well three different bias classes (Fig.~\ref{fig:orthoeffect} shows the extracted feature vectors at $\Gamma$). We encode the biases as three different colors (green, orange and blue), while the target class is represented by the arrows marker (triangle, square and circle). Typically, training a deep model without taking biases into account produces feature representations shown in Fig.~\ref{fig:orthoeffect}a: here, the loss on the target classes is minimized (three distinct groups are formed depending on the arrow marker), but it is driven by a heavy bias (the colors of the arrows). The purpose of EnD is to disentangle the representations belonging to the same bias class (color) and to entangle the representations with the same target class (the arrow's marker). Fig.~\ref{fig:orthoeffect}b represents the effect of EnD on the deep representations: while the disentangling term un-groups the biased example's representations, i.e. makes corresponding vectors almost orthogonal, the entangling one promotes correlations between samples having the same target. 

\subsection{Data correlations}
\label{sec:ddc}
Our main goal is to train our model to correctly classify the data into the $T$ possible classes, preventing the use of the bias features provided in the data. Towards this end, we aim at inserting an information bottleneck: the information related to these biases will be used as little as possible for the target classification task.\\
We can build a \emph{similarity matrix} $G \in \mathbb{R}^{M \times M}$:
\begin{equation}
    \label{eq:Gram}
    G = \left(\tilde{\boldsymbol{y}}\right)' \cdot \tilde{\boldsymbol{y}},
\end{equation}
where $(\cdot)'$ indicates transposed matrix and $ \tilde{\boldsymbol{y}}$ indicates a per-representation normalization
\begin{equation}
    \label{eq:norm}
    \tilde{\boldsymbol{y}}_{i} = \frac{\boldsymbol{y}_{i}}{\| \boldsymbol{y}_{i} \|_2} \forall i \in [1, M].
\end{equation}
Hence, every $g_{i, j}$ entry between two patterns $i,j$ in $G$ indicates their correlation:
\begin{equation}
    g_{i,j} = \left(\tilde{\boldsymbol{y}}_{i} \right)' \cdot \tilde{\boldsymbol{y}}_{j}.
\end{equation}
$G$ is a special case of \emph{Gramian matrix}, as any $g_{i,j}~\in~[-1;+1]$ and indicates the difference in the direction between any two $\boldsymbol{y}_{i}$ and $\boldsymbol{y}_{j}$. $G$ has some properties:
\begin{itemize}
    \item is a symmetric, positive semi-definite matrix;
    \item all the elements in the main diagonal are exactly $1$ by construction;
    \item if the subset of outputs $\tilde{\mathbf{y}}$ forms an ortho-normal basis (or $G$ is full-rank), then $G = \mathbb{I}$ by definition.
\end{itemize}
Handling these relations, we are going to build our regularization strategy, which consists in two terms:
\begin{itemize}
    \item a \emph{disentangling} term, whose task is to try to de-correlate as much as possible all the patterns belonging to the same bias class $b$;
    \item an \emph{entangling} term, which attempts to force correlations between data from different bias classes but having the same target class $t$.
\end{itemize}

\subsection{The EnD regularizer}
\label{sec:EnD}
The regularization $R$ we propose blends the disentangling $R_{\perp}$ and entangling $R_{\parallel}$ terms by setting
\begin{equation}
    R = \alpha R_{\perp} + \beta R_{\parallel},
\end{equation}
\noindent where $\alpha$ and $\beta$ are proper multipliers.
In the following, we are going to describe in detail the disentangling and the entangling terms.

\subsubsection{Disentangling term}
In order to disentangle biased representations, at training time, we select the patterns belonging to a bias class $b$ and 
build the corresponding Gramian matrix
\begin{equation}
    \label{eq:Grambias}
    G^{-,b} = \left(\tilde{\boldsymbol{y}}^{-,b}\right)' \cdot \tilde{\boldsymbol{y}}^{-,b}.
\end{equation}
Then, we enforce de-correlation between the features belonging to the same class: ideally, we would like to get $G^{-,b}~\rightarrow~\mathbb{I}~\forall~b$. To this end, we introduce the regularization term 
\begin{equation}
    \label{eq::strongregu}
    R_{\perp} = \frac{1}{B}\sum_{b=1}^B \frac{1}{\left(M^{-,b}\right)^2}\sum_{i, j} \left| g_{i,j}^{-,b} \right|
\end{equation}
\noindent that promotes minimization of the off-diagonal elements in $G^{-,b}$, $\forall b$.

\subsubsection{Entangling term}
While $R_{\perp}$ discourages the model to learn biases, the model should also build strong correlations between patterns belonging to different bias classes, but to the same target class $t$. With an orthogonal approach to the one used to derive \eqref{eq:Grambias}, we compute the Gramian matrix for the patterns belong to the same target class $t$:
\begin{equation}
\label{eq:Gramtarget}
    G^{t,-} = \left(\tilde{\boldsymbol{y}}^{t,-}\right)' \cdot \tilde{\boldsymbol{y}}^{t,-}.
\end{equation}
Let us focus, now, on the vector $\boldsymbol{g}_{i}^{t,-}$, extracted from the $i$-th column of $G^{t,-}$: it expresses how the $i$-th pattern correlates to all the other patterns which will be grouped to the same $t$-th target class. As a first option, we might ask the model to correlate the $i$-th pattern to all the other patterns having the same target class $t$, deriving the pattern entangling rule as the opposite of the disentangling rule in~\eqref{eq::strongregu}:
\begin{equation}
    \label{eq:dumbentangle}
    \hat{R}_{\parallel} = 1 - \frac{1}{T} \sum_{t=1}^T \frac{1}{\left(M^{t,-}\right)^2}\sum_{i, j} g_{i,j}^{t,-}
\end{equation}
In this formulation we are asking all the $g_{i,j}^{t,-}\rightarrow 1$, correlating the features as much as possible. However, \eqref{eq:dumbentangle} has a major shortcoming: it simply forces again correlations according to the target class $t$ regardless the bias information, which might be re-introduced. This is already done at a more general level by the loss function minimization as in~\eqref{eq:uprule}: it is desirable to have a term which entangles features having the same target class, but belonging to \emph{different} bias classes. Towards this end, we can re-write~\eqref{eq:dumbentangle} maximizing the correlations between each single example $\boldsymbol{y}_i$ and every other example $\boldsymbol{y}_j$ such that $\mathcal{T}(\boldsymbol{y}_i)=\mathcal{T}(\boldsymbol{y}_j)$ but, at the same time, $\mathcal{B}(\boldsymbol{y}_i)\neq \mathcal{B}(\boldsymbol{y}_j)$. Hence, our entangling term reads
\begin{align}
    R_{\parallel} = 1- \frac{1}{M} \sum_{i=1}^M & \frac{1}{\sum\limits_{b \neq \mathcal{B}(\boldsymbol{y}_i)} M^{\mathcal{T}(\boldsymbol{y}_i), b}}\cdot \nonumber\\
    \cdot&\sum_{j} \bar{\delta}\left[\mathcal{B}(\boldsymbol{y}_i), \mathcal{B}(\boldsymbol{y}_j)\right] \cdot {g}_{i,j}^{\mathcal{T}(\boldsymbol{y}_i), -} ,
\end{align}
where
\begin{equation}
    \bar{\delta}(a, b)=\left\{
    \begin{array}{ll}
        0 & a =b\\
        1 & a \neq b
    \end{array}
    \right . .
\end{equation}
\section{Experiments}
\label{sec:experiments}

In the experiments we present in this section, we aim to remove different types of biases such as color, age, gender which can have a high impact on classification performance when recognizing, for example, attributes such as hair color and presence of makeup on facial images. Additionally, we also show how this technique can help in sensitive tasks such as in the medical field, specifically in COVID-19 detection from CXR images. In all the results tables, the best results are denoted as boldface, the second best results are underlined. ``Vanilla'' denotes the baseline model performance for the learning problem, with no debiasing technique applied. All the EnD's results are averaged over three different runs.\footnote{The source code, written using PyTorch~1.7, 
will be made publicly available in the final version of the article. The hyperparameters used for the proposed experiments are optimized using a validation set or k-folding cross-validation depending on the dataset.}

\subsection{Controlled experiments}

In this section we describe the controlled experiments that we performed in order to assess the performance of \absname. Full control over the amount and type of bias allows to correctly analyze \absname's behavior, excluding noise and uncertainty given by real-world data.

\subsubsection{Biased MNIST}

\begin{figure}
    \centering
    \includegraphics[width=\columnwidth]{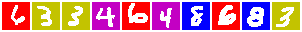}
    \caption{\textbf{Biased MNIST} by Bahng~\emph{et~al.}~\cite{bahng2019rebias}, where the background colors highly correlate with the digit classes.}
    \label{fig:biased-mnist}
\end{figure}

\newcommand{\accnnn}{52.30}

\begin{table}
    \centering
    \begin{tabular}{l c c c c}
        \toprule
        \multirow{2}{*}{Method}        &\multicolumn{4}{c}{$\rho$ values}\\
         & 0.999 & 0.997 & 0.995 & 0.990 \\ 
        \midrule
        Vanilla                         & 10.4              & 33.4              & 72.1              & 89.1\\
        HEX~\cite{wang2018learning}     & 10.8              & 16.6              & 19.7              & 24.7\\
        LearnedMixin~\cite{ClarkYZ19}   & 12.1              & 50.2              & 78.2              & 88.3\\
        RUBi~\cite{cadene2019rubi}      & 13.7              & 43.0              &\underline{90.4}   &\underline{93.6}\\
        ReBias~\cite{bahng2019rebias}   & \underline{22.7}  & \underline{64.2}  & 76.0              &88.1\\
        EnD                             & \textbf{\accnnn}    & \textbf{83.70}    & \textbf{93.92}    & \textbf{96.02}\\
                                        & $\pm$ 2.39       & $\pm$ 1.03        & $\pm$ 0.35        & $\pm$ 0.08\\
        \bottomrule
    \end{tabular}
    \caption{\textbf{Biased MNIST performance on the unbiased test set.}}
    \label{table:mnist-results}
\end{table}

\begin{figure*}
    \begin{subfigure}{\columnwidth}
        \centering
        \includegraphics[width=\columnwidth]{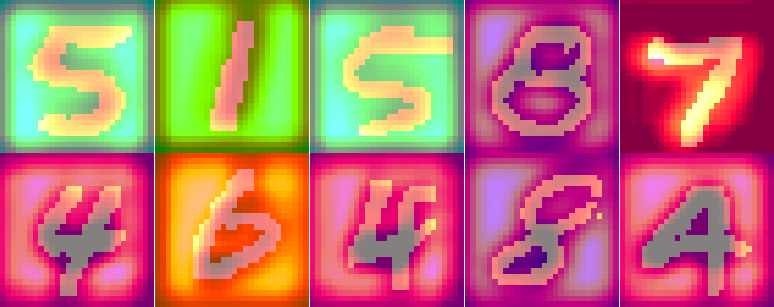}
        \caption{~}
        \label{fig:cam-biased}
    \end{subfigure}
    \hfill
    \begin{subfigure}{\columnwidth}
            \centering
            \includegraphics[width=\columnwidth]{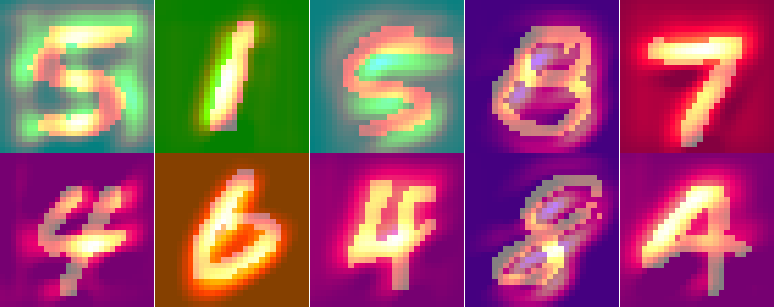}
            \caption{~}
            \label{fig:cam-unbiased}
    \end{subfigure}
    \caption{\textbf{Grad-CAM on Colored~MNIST:}~\cite{selvaraju2017grad} vanilla model (a) and \absname-regularized model~(b).}
    \label{fig:gradcam-mnist}
\end{figure*}

\begin{figure*}
    \begin{subfigure}{0.49\columnwidth}
        \centering
        \includegraphics[width=\columnwidth]{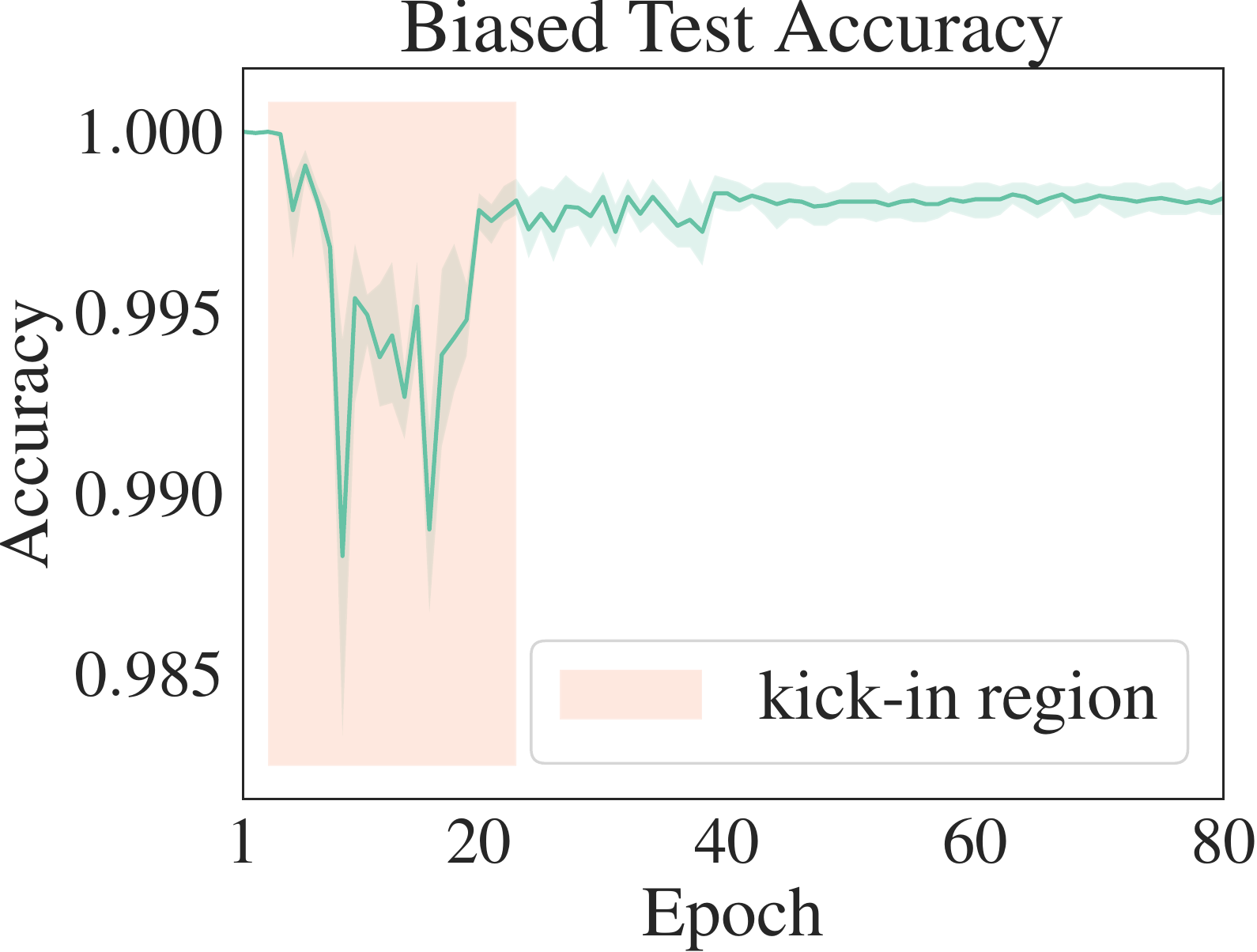}
        \caption{~}
        \label{fig:mnist-acc-biased}
    \end{subfigure}
    \begin{subfigure}{0.49\columnwidth}
            \centering
            \includegraphics[width=\columnwidth]{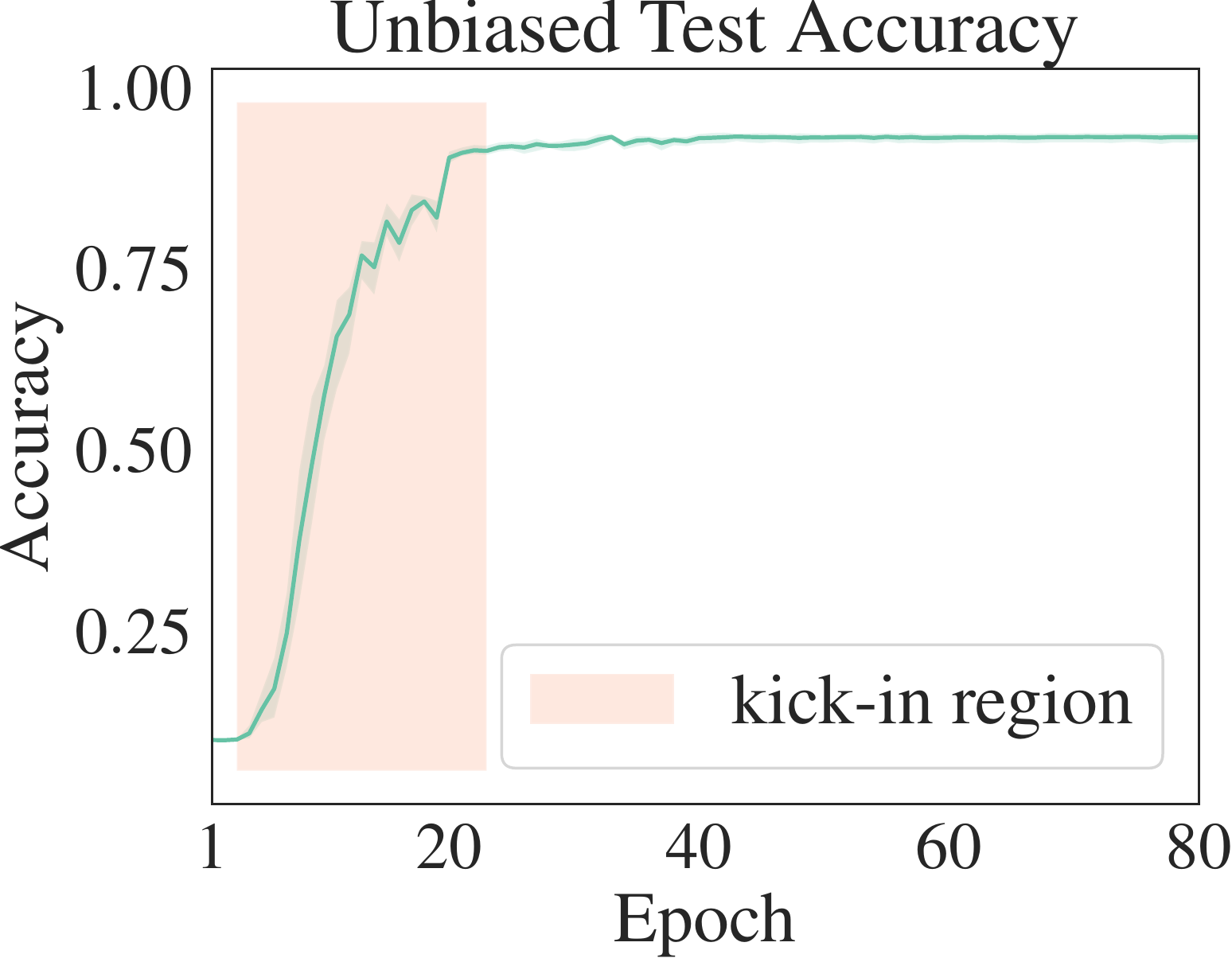}
            \caption{~}
            \label{fig:mnist-acc-unbiased}
    \end{subfigure}
    \begin{subfigure}{0.49\columnwidth}
        \centering
        \includegraphics[width=\columnwidth]{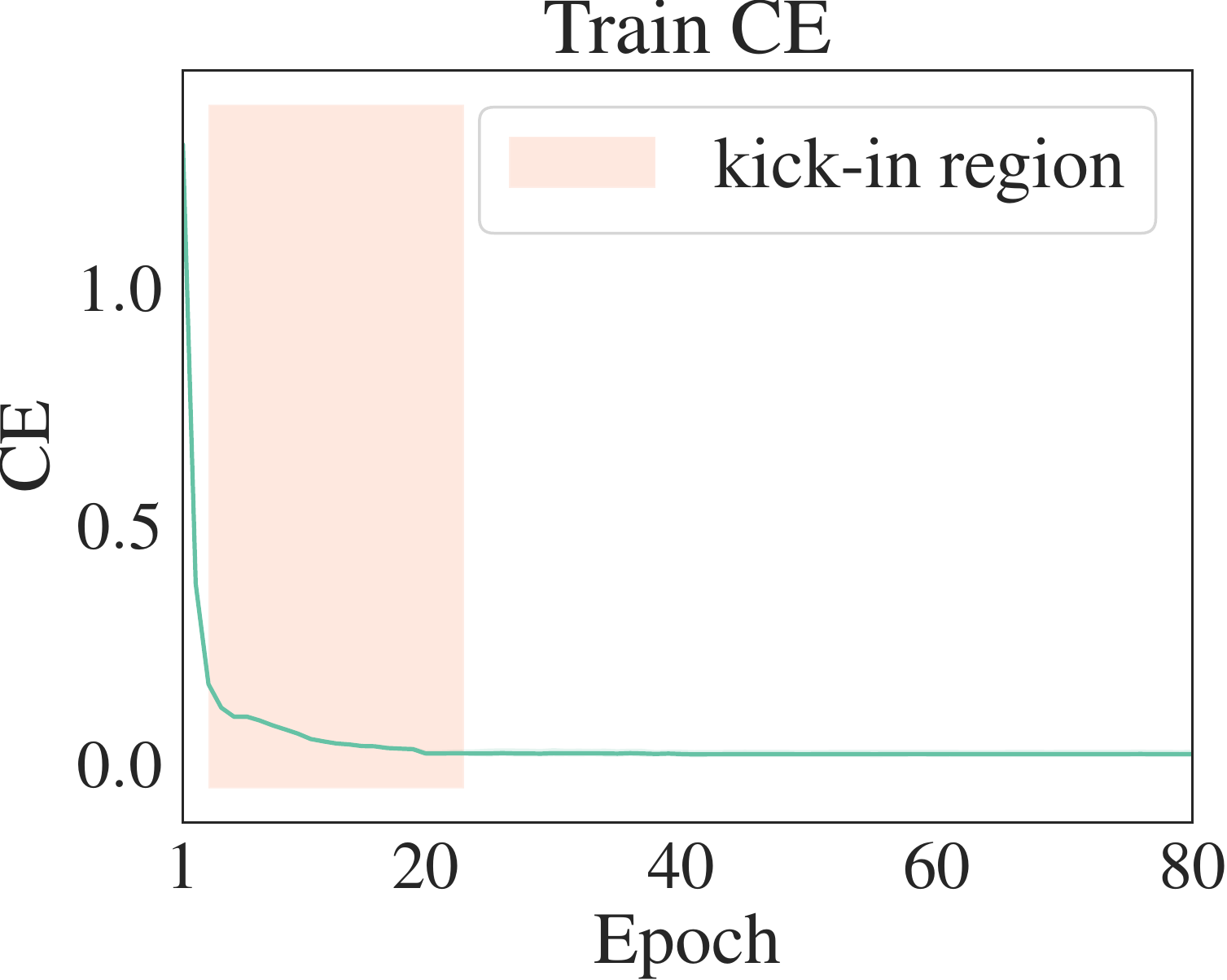}
        \caption{~}
        \label{fig:mnist-train-bce}
    \end{subfigure}
    \begin{subfigure}{0.49\columnwidth}
            \centering
            \includegraphics[width=\columnwidth]{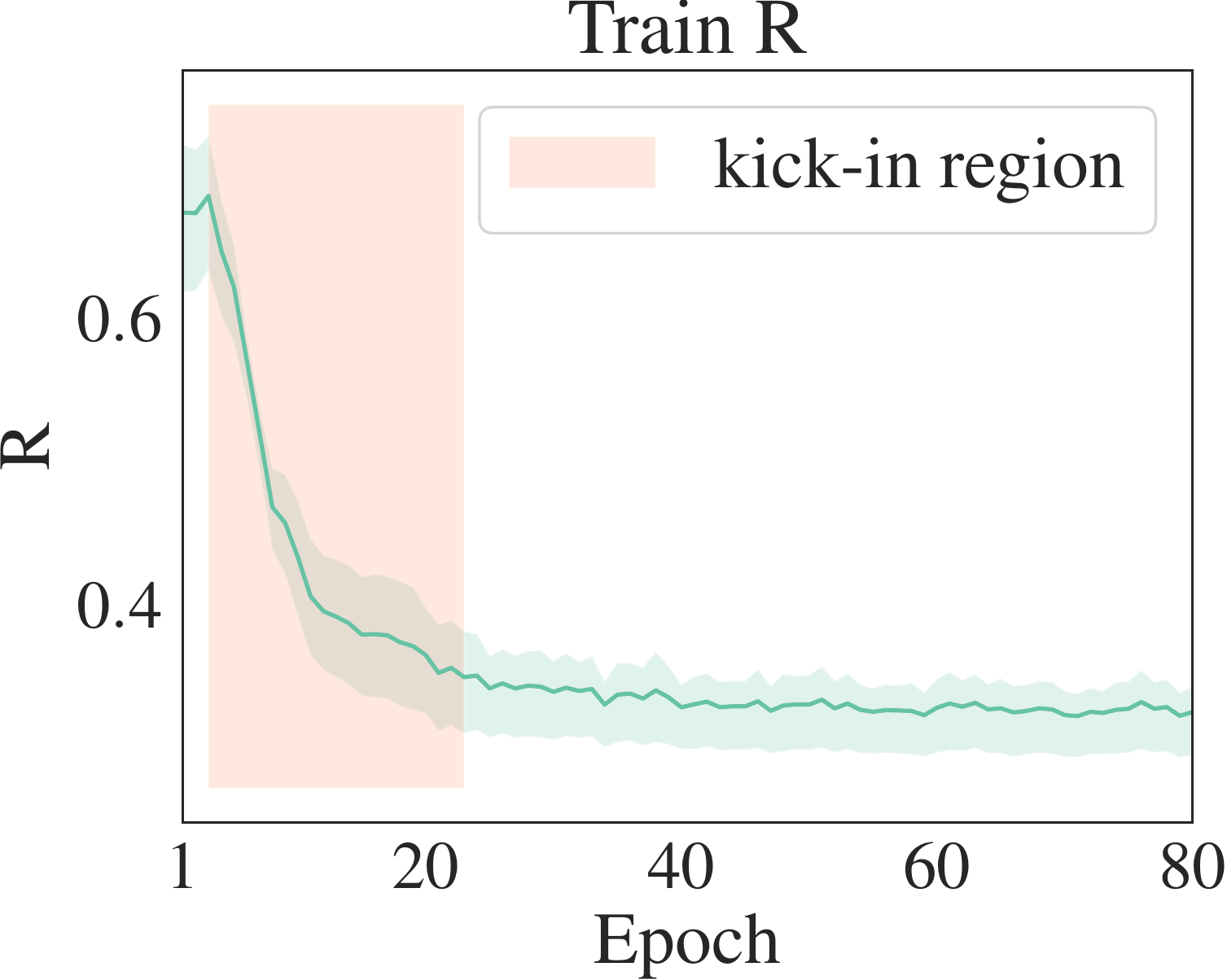}
            \caption{~}
            \label{fig:mnist-train-abs}
    \end{subfigure}
    \caption{\textbf{\absname learning curves on Colored~MNIST for $\boldsymbol{\rho}\mathbf{=0.995}$.} Biased accuracy~(a), unbiased accuracy~(b), $L$ value on the training set~(c) and  $R$ value on the training set~(d).}
    \label{fig:learning-curves}
\end{figure*}

We test our method on a synthetic dataset, where we can control the bias in the training data. We use the \emph{Biased MNIST} dataset proposed by Bahng~\emph{et~al.}~\cite{bahng2019rebias}. This dataset is constructed from the MNIST dataset~\cite{lecun2010mnist} by injecting a color into the images background, as shown in Figure~\ref{fig:biased-mnist}. Each digit 
is associated with one of ten pre-defined colors. 
To assign the color bias to an image of a given target class, 
the pre-defined color 
is selected with a probability $\rho$, and any other color is chosen with a probability $(1-\rho)$. To vary the level of difficulty in the dataset, the authors select $\rho \in \{0.990, 0.995, 0.997, 0.999 \}$. Higher values of $\rho$ correspond to higher correlation between target class and bias class (color). Two testing datasets are constructed with the same criterion: \emph{biased}, with $\rho = 1.0$, and \emph{unbiased}, with $\rho = 0.1$. Given the low correlation between color and digit class in the unbiased test set, models must learn to classify shapes instead of colors in order to reach a high accuracy.

\noindent \textbf{Setup.} We use the network architecture proposed by Bahng~\emph{et~al.}~\cite{bahng2019rebias}, consisting of four convolutional layers with $7\times 7$ kernels. The \absname regularization term is applied on the average pooling layer, before the fully connected classifier of the network.

\noindent \textbf{Results.} Results are shown in Table~\ref{table:mnist-results}. \absname's results are averaged across three different runs for each value of $\rho$. For all values of $\rho$ we report the accuracy obtained by \absname on the unbiased evaluation set, compared with other debiasing algorithms. 

\noindent \textbf{\absname successfully mitigates bias propagation.} The improvement obtained with \absname with respect to the baseline model is noticeable, especially in the higher levels of difficulty. We observe an increase of accuracy across all values of $\rho$. Notably, for $\rho=0.999$ the vanilla model reaches 10.4\% accuracy, meaning that the background color is used as the only cue for classifying the digits, whereas employing \absname yields an accuracy of \accnnn\%. Figure~\ref{fig:gradcam-mnist} shows the effect of \absname, using Grad-CAM~\cite{selvaraju2017grad} to highlight the important regions of the input image for the model prediction. We observe that the vanilla model (Figure~\ref{fig:cam-biased}) focuses on the background, while the \absname-regularized model (Figure~\ref{fig:cam-unbiased}) correctly learns to focus on the digit shape.

\noindent \textbf{Comparison with other techniques.} We observe that \absname yields the highest results among all of the compared debiasing algorithms. Such gap is especially higher in the most difficult settings for $\rho \in \{0.999, 0.997\}$ where many algorithms are unable to generalize to the unbiased set, especially HEX~\cite{wang2018learning} and LearnedMixin~\cite{ClarkYZ19}. Some of the compared algorithms even show a collapse in accuracy compared to the vanilla baseline in certain cases (HEX for most values of $\rho$, LearnedMixin and ReBias for $\rho=0.990$).

\noindent \textbf{Ablation study.} We also perform an ablation study of \absname to analyze how each of the \absname's terms affect the performance of the trained model. For a fixed $\rho = 0.997$, we evaluate only the contribution of the disentangling term $R_{\perp}$ and disable the entangling term $R_{\parallel}$ by setting $\beta=0$. We then perform the opposite evaluation by setting $\alpha=0$, to only take into account the entangling term.
\begin{table}
    \centering
    \begin{tabular}{l c c c}
        \toprule
        \multirow{2}{*}{Setting} & \multirow{2}{*}{$\alpha$}   & \multirow{2}{*}{$\beta$}            & Unbiased \\
                &&&accuracy\\
        \midrule
        Vanilla & 0 & 0 & 33.4 \\
        Disentangling only & $[0; 1]$ & 0                  &  45.67 $\pm$ 0.67 \\
        Entangling only    & 0      & $[0; 1]$ &  75.36 $\pm$ 0.94 \\
        \midrule
        EnD                 & $[0; 1]$ & $[0; 1]$ & \textbf{83.70} $\pm$ 1.03 \\
        \bottomrule
    \end{tabular}
    \caption{\textbf{Ablation study of EnD on the Biased~MNIST dataset}, $\rho=0.997$.}
    \label{tab:abs-ablation}
\end{table}
The results are shown in Table~\ref{tab:abs-ablation}.
We observe that both the regularization terms contribute to boost the model's generalization capability. As expected, the best results are achieved when both of them are jointly applied. The entangling term yields a higher increase in performance compared to the disentangling one, however it is in general not always applicable, for example when, given some $i$-th sample $\boldsymbol{y}_i$,
\begin{equation}
    \nexists j~
    |~\mathcal{T}(\boldsymbol{y}_i)=\mathcal{T}(\boldsymbol{y}_j) \land \mathcal{B}(\boldsymbol{y}_i)\neq \mathcal{B}(\boldsymbol{y}_j)~\forall i.\nonumber
\end{equation}
The disentangling term provides a smaller benefit in this case, but, on the other hand, it can always be applied. We find that the ideal case for \absname is when both of the terms can be used in the learning process, leading to better generalization capabilities.
Furthermore, we observe a similar pattern in the learning process when employing the full \absname regularization for different values of $\rho$. Figure~\ref{fig:learning-curves} shows the learning curves for $\rho=0.995$. We notice how models tend to quickly learn the color bias in the first few epochs, as the accuracy on the biased test set is close to 100\%~(Figure~\ref{fig:mnist-acc-biased}). However, once the value of the loss (in this case, we have used the cross-entropy loss, Figure~\ref{fig:mnist-train-bce}) falls below a certain threshold, the contribution $R$ of the \absname term becomes predominant~(Figure~\ref{fig:mnist-train-abs}). In this phase, which we call \mbox{\emph{kick-in region}}, the optimization process begin to rapidly minimize $R$, stopping the model from relying on the bias-related features. This can be observed in the rapid increase of the accuracy on the unbiased test set~(Figure~\ref{fig:mnist-acc-unbiased}), whereas the biased accuracy momentarily drops as the models shift their focus from the background color to the digit shape.

\subsection{Real world datasets}
After benchmarking \absname in a controlled scenario on synthetic data, we move to real world datasets where biases might be subtle and harder to handle. In this section we aim at removing age and gender bias in different datasets. We also apply \absname on a computer-aided diagnosis task, where hidden biases might lead to sub-optimal generalization of the model.

\noindent \textbf{Setup.} For CelebA and IMDB Face, we use the ResNet-18 model proposed by He~\emph{et~al.}~\cite{he2016deep}. The network was pre-trained on ImageNet~\cite{imagenet_cvpr09}, except for the last fully connected layer. The \absname regularization is applied on the average pooling layer, before the fully connected classifier.
For \corda,\footnote{This dataset's name and the involved institutions are kept anonymous (just) in the reviewing process since it has not been publicly released yet.} we use a \mbox{DenseNet-121~\cite{huang2017densely}} encoder pre-trained on publicly available CXR data, which is then followed by a two-layer fully connected classifier.

\subsubsection{CelebA}
CelebA~\cite{liu2015faceattributes} is a dataset of for face-recognition tasks, providing 40 attributes for every image. Following Nam~\emph{et~al.}~\cite{nam2020learning}, we select \emph{BlondHair} and \emph{HeavyMakeup} as target attributes $t$ and \emph{Male} as bias attribute $b$. This choice is dictated by the fact that there is a high correlation between the target and the bias attributes (i.e. most women have blond hair or wear heavy makeup in this dataset). The dataset contains a total of 202,599 images, and following the official train-validation split we obtain 162,770 images for training and 19,867 images for testing our models. Nam~\emph{et~al.}~\cite{nam2020learning} build two types of testing dataset: \emph{unbiased}, by selecting the same number of samples for every possible value of the pair $(t,b)$, and \emph{bias-conflicting}, by removing from the unbiased set all of the samples where $b$ and $t$ are equal.

%
\begin{table}
    \centering
    \begin{tabular}{@{}l@{\hspace{0.5\tabcolsep}} l@{\hspace{0.5\tabcolsep}} l c c}
        \toprule
        && Method & Unbiased & Bias-conflicting\\
        \midrule
        \multirow{4}{*}{\rotatebox{90}{Learn}} & \multirow{4}{*}{\rotatebox{90}{HairColor}}
        &Vanilla                                         & 70.25 $\pm$ 0.35 & 52.52 $\pm$ 0.19 \\
        &&Group DRO~\cite{sagawa2019distributionally}     & \underline{85.43} $\pm$ 0.53 & \underline{83.40} $\pm$ 0.67\\
        &&LfF\cite{nam2020learning}                       & 84.24 $\pm$ 0.37 & 81.24 $\pm$ 1.38 \\
        &&\absname                                        & \textbf{91.21} $\pm$ 0.22 & \textbf{87.45} $\pm$ 1.06\\
        \midrule
        \addlinespace[1.0ex]
        \multirow{5}{*}{\rotatebox{90}{Learn}} & \multirow{4}{*}{\rotatebox{90}{\small HeavyMakeup}}
        &Vanilla                                     & 62.00 $\pm$ 0.02 & 33.75 $\pm$ 0.28 \\\addlinespace[0.4ex]
        &&Group DRO~\cite{sagawa2019distributionally} &64.88 $\pm$ 0.42 & \underline{50.24} $\pm$ 0.68 \\\addlinespace[0.4ex]
        &&LfF\cite{nam2020learning}                   & \underline{66.20} $\pm$ 1.21 & 45.48 $\pm$ 4.33 \\\addlinespace[0.4ex]
        &&\absname                                    & \textbf{75.93} $\pm$ 1.31 & \textbf{53.70} $\pm$ 5.24  \\\addlinespace[0.8ex]
        \bottomrule
    \end{tabular}
    \caption{\textbf{Performance on CelebA.}}
    \label{table:celeba-results}
\end{table}
\noindent \textbf{Results.} Following Nam~\emph{et~al.}~\cite{nam2020learning}, the accuracy is computed as average accuracy over all the $(t, b)$ pairs. 
Table~\ref{table:celeba-results} shows the results obtained on the CelebA dataset. We observe how the vanilla model heavily relies on the bias attribute, scoring a low accuracy especially on the bias-conflicting sets. EnD, on the other hand, outperforms the baseline in both the tasks. 
We report reference results~\cite{nam2020learning} of other debiasing algorithms, specifically Group~DRO~\cite{sagawa2019distributionally} and LfF~\cite{nam2020learning}, for comparison with \absname. The results we obtain are significantly higher across most of the evaluation sets, and comparable with Group DRO and LfF on the bias-conflicting set when the target attribute is HeavyMakeup. 

\subsubsection{IMDB Face}

\begin{figure}
    \centering
    \begin{subfigure}{\columnwidth}
        \centering
        \includegraphics[width=\columnwidth]{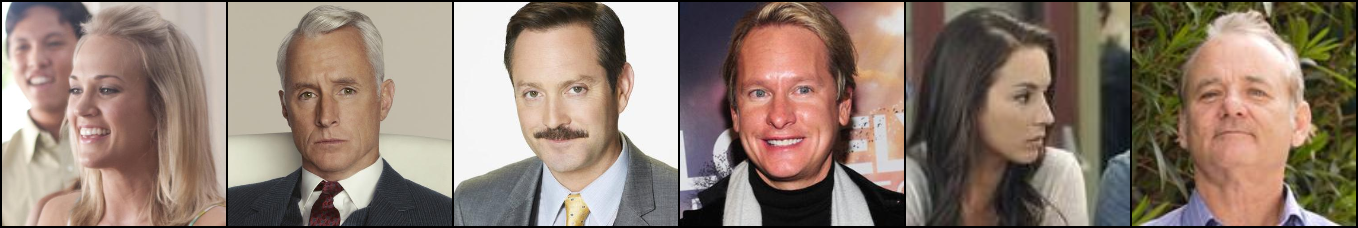}
        \caption{~}
        \label{fig:imdb-eb1}
    \end{subfigure}
    \hfill
    \begin{subfigure}{\columnwidth}
            \centering
            \includegraphics[width=\columnwidth]{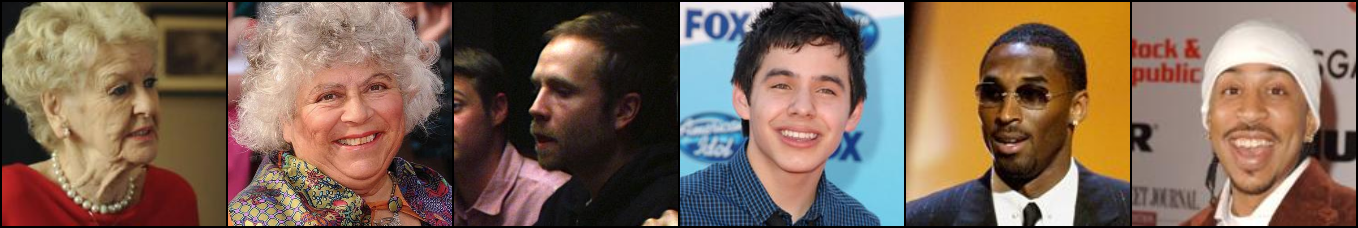}
            \caption{~}
            \label{fig:imdb-eb2}
    \end{subfigure}
    \caption{\textbf{IMDB train splits:} EB1 (a) and EB2 (b).}
    \label{fig:imdb-eb}
\end{figure}

The IMDB Face dataset~\cite{Rothe-IJCV-2018} contains 460,723 face images annotated with age and gender information. To filter out the misannotated labels of this dataset~\cite{Rothe-IJCV-2018, torralba2011unbiased}, Kim~\emph{et~al.}~\cite{Kim_2019_CVPR} use a model trained on the Audience benchmark~\cite{eidinger2014age}, keeping the images where the prediction matches the provided label. Following Kim~\emph{et~al.}'s proposed data split, 20\% of the IMDB is used as test set, containing samples with age 0-29 or 40+. The remaining data is then split into two extreme-bias subset: \emph{EB1} contains women in the age range 0-29 and men with age 40+, while \emph{EB2} contains men aged 0-29 and women 40+. Thus, when learning to predict the gender attribute, the bias is given by the age and vice-versa. An example of the EB1 and EB2 training sets is shown in Figure~\ref{fig:imdb-eb}.

%
\begin{table}
    \centering
    \begin{tabular}{@{}l@{\hspace{0.8\tabcolsep}} l c c c c}
        \toprule
        &\multirow{2}{*}{Method}& \multicolumn{2}{c}{Trained on EB1} & \multicolumn{2}{c}{Trained on EB2} \\
        & & EB2 & Test & EB1 & Test \\
        \midrule
        \multirow{5}{*}{\rotatebox{90}{Learn Gender}} 
        & Vanilla                                 & 59.86             & 84.42             & 57.84             & 69.75 \\
        &BlindEye~\cite{alvi2018turning}         & 63.74             & 85.56             & 57.33             & 69.90 \\
        &Kim~\emph{et~al.}~\cite{Kim_2019_CVPR}  & \textbf{68.00}    & \underline{86.66} & \underline{64.18} & \underline{74.50} \\
        &EnD                    & \underline{65.49} & \textbf{87.15}    & \textbf{69.40}    & \textbf{78.19} \\
        &                                        & $\pm$ 0.81        & $\pm$ 0.31        & $\pm$ 2.01        &$\pm$ 1.18 \\
        \midrule
        \multirow{5}{*}{\rotatebox{90}{Learn Age}} 
        &Vanilla                               & 54.30            & 77.17           & 48.91           & 61.97 \\
        &BlindEye~\cite{alvi2018turning}       & \underline{66.80}& 75.13           &\underline{64.16}& 62.40 \\
        &Kim~\emph{et~al.}~\cite{Kim_2019_CVPR}& 65.27            &\underline{77.43}& 62.18           &\underline{63.04} \\
        &EnD                                   & \textbf{76.04}   & \textbf{80.15}  & \textbf{74.25}  & \textbf{78.80} \\  
        &                                      & $\pm$ 0.25       & $\pm$ 0.96      & $\pm$ 2.26      &$\pm$ 1.48 \\
        \bottomrule
    \end{tabular}
    \caption{\textbf{Performance on IMDB Face.} When gender is learned, age is the bias, and when age is learned the gender is the bias.}
    \label{table:imdb-results}
\end{table}

\noindent \textbf{Results.} Table~\ref{table:imdb-results} shows the results obtained on the IMDB Face dataset. 
We performed two main experiments: gender and age prediction. Besides the perfomance evaluation on the test set, when training on EB1 we also tested the model's performance on EB2, and viceversa. This allows us to better evaluate the 
bias features' influence on the model prediction. We notice how the baseline model is heavily biased towards age when predicting gender, and towards gender when predicting age. This can be observed on the performance achieved on the EB2 and EB1 sets, both for gender and age prediction. 
When employing our regularization term, we observe an increase across all of the obtained results: in particular, when training on EB2 for age prediction, we notice an increase from 48.91\% to 74.25\% on the EB1 set. We also report reference results of other debiasing algorithms, specifically BlindEye~\cite{alvi2018turning} and the adversarial approach proposed by Kim~\emph{et~al.}~\cite{Kim_2019_CVPR}. In general, \absname obtains the best results among all the other debiasing algorithms we compared to.

\subsubsection{COVID CXR dataset}
\corda is a dataset comprising 898 Chest X-Ray images (CXR) that were collected during March and April 2020 by radiology units at \cdss and \slg, in \country. Virus testing (nasopharingeal swab) was used to determine the presence or absence of COVID-19 infection. The dataset can be split by collecting institution, resulting in \cordacdss with 297 images of COVID-19 positive patients and 150 of negative ones, and \cordaslg with 129 positives and 322 negatives. Recent literature~\cite{Cruz2020OnTC, maguolo2020critic, tartaglione2020unveiling} shows that merging CXRs coming from different sources poses bias issues, since differences in acquisition techniques given by the scan machines or composition of the population sample might be used by the deep model to distinguish the provenance of the data itself, even when pre-processing techniques are employed. For \corda, we notice that data coming from \cdss contain a majority of positive samples, while data coming from \slg have a majority of negative samples. Hence, if distinguishing features are embedded in the scans, then the networks might learn to discriminate the source of the data, instead of actually classifying between COVID positives and negatives.
To build the test sets, we use 30\% of \cordacdss and 30\% of \cordaslg. The remaining data are then merged and used as training set. Testing on the two distinguished sets allows us to assess whether the prediction of the models are biased towards the origin of the data.

\begin{table}
    \centering
    \begin{tabular}{l c c c }
            \toprule
            & \multicolumn{3}{c}{Test on \cordacdss} \\
            \cmidrule{2-4}
            & TPR & TNR & BA \\
            \midrule
            
            Vanilla & 69.99 $\pm$ 3.27 & 59.26 $\pm$ 2.09 & 64.63 $\pm$ 2.50 \\
            \absname & 68.16 $\pm$ 2.08 & 76.30 $\pm$ 2.10 & \textbf{72.22} $\pm$ 0.01 \\ 
            \midrule
            
            & \multicolumn{3}{c}{Test on \cordaslg}  \\
            \cmidrule{2-4}
            & TPR & TNR & BA \\
            \midrule
            Vanilla &  52.14 $\pm$ 3.20 & 87.63 $\pm$ 4.37 &  69.88 $\pm$ 2.95 \\
            \absname & 68.37 $\pm$ 6.04 & 84.51 $\pm$ 3.04 & \textbf{75.94} $\pm$ 1.62 \\
            \bottomrule
        \end{tabular}
    \caption{\textbf{Performance on \corda}, sorted by collecting institution.}
    \label{table:cxr-subset}
\end{table}
\noindent \textbf{Results.} The results obtained on \cordacdss and \cordaslg are presented in Table~\ref{table:cxr-subset}.
We observe how the vanilla model is in fact biased towards the source of the data. On \cordacdss (which contains mostly positive samples) the vanilla model shows a higher true positive rate (TPR) and a lower true negative rate (TNR). On the other hand, on \cordaslg (which contains mostly negative samples) we notice a lower TPR compared to the sensibly higher TNR. 
Employing \absname helps in improving the results in this case too. While maintaining a similar TPR on \cordacdss and TNR on \cordaslg, we obtain an improvement of the TNR 59.26\%$\rightarrow$76.30\% and of the TPR 52.14\%$\rightarrow$68.37\% on \cordacdss and \cordaslg, respectively. This also results in an increased balanced accuracy (BA) on both of the test sets.
\begin{figure}[h]
    \centering
    \begin{subfigure}{0.49\columnwidth}
            \centering
            \includegraphics[width=\columnwidth]{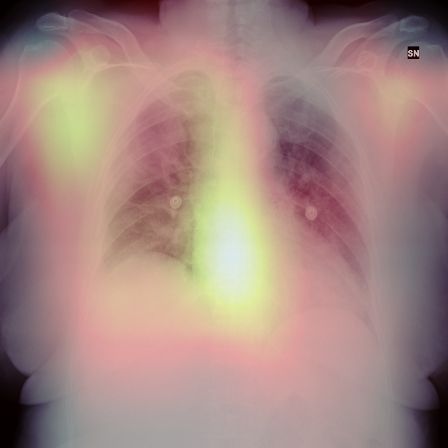}
            \caption{~}
            \label{fig:mean-biased}
    \end{subfigure}
    \hfill
    \begin{subfigure}{0.49\columnwidth}
            \centering
            \includegraphics[width=\columnwidth]{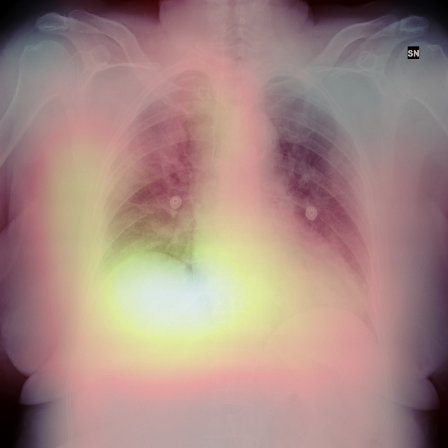}
            \caption{~}
            \label{fig:mean-unbiased}
    \end{subfigure}
    
    \caption{\textbf{Grad-CAM on \corda}: vanilla model (a) and EnD-regularized model (b).}
    \label{fig:gradcam-cxr}
\end{figure}
As a further insight, we observe in Figure~\ref{fig:mean-biased} that the vanilla model 
focuses on irrelevant regions outside the lungs area, while the EnD-regularized model mainly focuses on the lower lobes of the lungs~(Figure~\ref{fig:mean-unbiased}).

\section{Conclusion}
\label{sec:conclusion}

In this work we aimed at EnD-ing the selection of biased features in deep model trained on biased datasets. Towards this end, we have designed a regularizer whose task is to either disentangle deep feature representations with the same bias and to entangle deep features with different biases, but belonging to the same target classification class. 
Differently from other de-biasing techniques, we do not introduce any additional parameters to be learned and we do not modify the input data: the model itself is naturally driven into choosing deep features which are unbiased, without introducing additional priors to the data. Our experiments show the effectiveness of EnD when compared to other state-of-the-art techniques, excelling in the cases of heavily-biased data (like $\rho=0.999$ for Biased~MNIST or IMDB). 
As an application case, we have also tested the effect of EnD on the COVID diagnosis from CXR images, where the bias is given by the data source and it is not straightforward to detect. 
In this case we have observed an overall improvement of the performance on the test set as well, showing that our technique may be employed to build more reliable models even in more sensitive tasks.
\newpage
\bibliographystyle{ieee_fullname}
\bibliography{main}

\end{document}